\pdfoutput=1
\documentclass[11pt]{article}

\usepackage[final]{acl}

\usepackage{times}
\usepackage{latexsym}
\usepackage{array}
\usepackage[utf8]{inputenc}

\usepackage{microtype}

\usepackage{inconsolata}

\usepackage{graphicx}

\usepackage{amsmath}
\usepackage{amsfonts}
\usepackage{algorithm}
\usepackage{algpseudocode}
\usepackage{mdwlist}
\usepackage{multirow}
\usepackage{subcaption}
\usepackage{float}
\usepackage{epigraph}
\usepackage{booktabs}
\usepackage{tabularray}
\usepackage{adjustbox}
\usepackage{multirow}
\usepackage{multicol}
\usepackage{wrapfig}
\usepackage{scalerel}
\usepackage{float}

\usepackage[T1]{fontenc}    % use 8-bit T1 fonts

\usepackage[utf8]{inputenc}

\usepackage{microtype}

\usepackage{inconsolata}

\usepackage[normalem]{ulem} % underlines
\usepackage{soul}
\usepackage{times}
\usepackage{latexsym}
\usepackage{datetime}       % dates
\usepackage{booktabs}       % professional-quality tables
\usepackage{nicefrac}       % compact symbols for 1/2, etc.
\usepackage{url}            % simple URL typesetting
\usepackage{fnpct}           % nicer footnotes
\usepackage{enumitem}        % nicer itemize
\usepackage{tcolorbox}       % colored background
\usepackage{csquotes}        % quotation marks
\usepackage{xspace}          % adds whitespace after macros
\usepackage{graphicx}

\makeatletter
\xspaceaddexceptions{\csq@qclose@i}
\makeatletter

\usepackage[scaled=0.85]{helvet}

\usepackage{bbm}
\usepackage{amsfonts}
\usepackage{amssymb}
\usepackage{mathtools}
\usepackage{amsmath}
\usepackage{bm}
\usepackage{amsthm}
\usepackage{thmtools} 
\usepackage{thm-restate}
\usepackage{cancel}

\usepackage{nicefrac}

\usepackage[createShortEnv,commandRef=Cref]{proof-at-the-end}

\newtheoremstyle{resultstyle} % name of the style to be used
  {.8em} % space above
  {.8em} % space below
  {} % font in the body of the theorem
  {} % indent amount
  {\bfseries} % theorem head font
  {.} %  punctuation after theorem head
  {.5em} % space after theorem head
  {\thmname{#1 }\thmnumber{#2. }\textbf{\thmnote{#3}}}

\theoremstyle{resultstyle}

\usepackage{algorithm}
\usepackage{algpseudocode}  % requires algorithmicx
\makeatletter
\algnewcommand{\LineComment}[1]{\Statex \hskip \ALG@thistlm \textcolor{blue}{// #1}}
\algnewcommand{\FirstLineComment}[1]{\Statex \hskip\ALG@tlm \textcolor{blue}{\(\triangleright\) #1}}
\algnewcommand{\InlineComment}[1]{\hfill\textcolor{blue}{\(\triangleright\) #1}}
\makeatother

\usepackage{cleveref}
\crefname{section}{\S}{\S\S}
\Crefname{section}{\S}{\S\S}
\crefformat{section}{\S#2#1#3}
\crefname{figure}{Fig.}{Fig.}
\crefname{alg}{Alg.}{Alg.}
\crefname{line}{line}{lines}
\crefname{appendix}{App.}{App.}
\crefname{equation}{eq.}{eqs.}
\crefname{table}{Table}{Tables}
\crefname{proposition}{Proposition}{Propositions}
\crefname{assumption}{Assump.}{Assumps.}
\crefname{lemma}{Lemma}{Lemmas}
\crefname{definition}{Defn.}{Defns.}
\crefname{hypothesis}{Hypothesis}{Hypotheses}
\crefname{estimator}{Estimator}{Estimators}
\crefname{theorem}{Theorem}{Theorems}
\crefname{thm}{Theorem}{Theorems}
\crefname{result}{Result}{Results}

\usepackage{cellspace}
\newcommand\cincludegraphics[2][]{\raisebox{-0.4\height}{\includegraphics[#1]{#2}}}
\usepackage{setspace}

\usepackage{xcolor}
\definecolor{accentblue}{RGB}{51,102,153}
\definecolor{accentorange}{RGB}{204,102,0}
\definecolor{accentgray}{RGB}{100,100,100}
\definecolor{lightblue}{RGB}{230,240,250}
\definecolor{lightorange}{RGB}{255,245,230}

\definecolor{mamlcolor}{RGB}{51,102,153}        % Blue: MAML/meta-learning
\definecolor{vanillacolor}{RGB}{100,100,100}    % Gray: vanilla baseline
\definecolor{phasecolor}{RGB}{153,51,102}       % Purple: phase transitions
\definecolor{improvementcolor}{RGB}{34,139,34}  % Green: improvements/gains
\definecolor{degradationcolor}{RGB}{204,51,51}  % Red: degradations/losses
\definecolor{dynamicscolor}{RGB}{204,102,0}     % Orange: learning dynamics

\definecolor{lightmaml}{RGB}{230,240,250}
\definecolor{lightphase}{RGB}{245,230,245}

\algrenewcommand{\alglinenumber}[1]{\color{accentgray}\tiny #1}

\title{Learning Dynamics of Meta-Learning in Small Model Pretraining}

\author{
    \textbf{
        David Demitri Africa\thanks{Corresponding author: \textbf{david.demitri.africa@gmail.com}} ~~~
        Yuval Weiss
        } \\
    \textbf{
         Paula Buttery
         ~~~
         Richard Diehl Martinez
         }\\
    University of Cambridge
}

\begin{document}
\maketitle
\begin{abstract}
Large language models are powerful but costly. We ask whether meta-learning can make the pretraining of small language models not only faster but also more interpretable. We integrate first–order MAML with subset-masked LM pretraining, producing four LLama-style decoder-only models (11M–570M params), and evaluate on multilingual Universal NER. Compared with vanilla training, our hybrid setup (i) reaches the same loss up to 1.6× sooner, (ii) yields modest but consistent average gains on Universal NER at medium/large scales under equal compute (+2–3 percentage points), and (iii) and (iii) reveals phase-like learning dynamics: models first diversify their representations, then compress them in a pattern that aligns with improved episodic accuracy. These observations are correlational, not causal, and we do not claim generality beyond NER or across seeds. We also document a trade-off: perplexity on Paloma (a diverse language modeling benchmark spanning 18 domains; \citet{magnusson2024paloma}) is worse at most scales. Code, checkpoints and analysis logs are released.

% tweak row / column spacing (optional)
\renewcommand{\arraystretch}{1.1}   % vertical padding
\setlength{\tabcolsep}{4pt}         % horizontal padding

\begin{tabular}{@{}c l@{}}
  \cincludegraphics[width=1.9em]{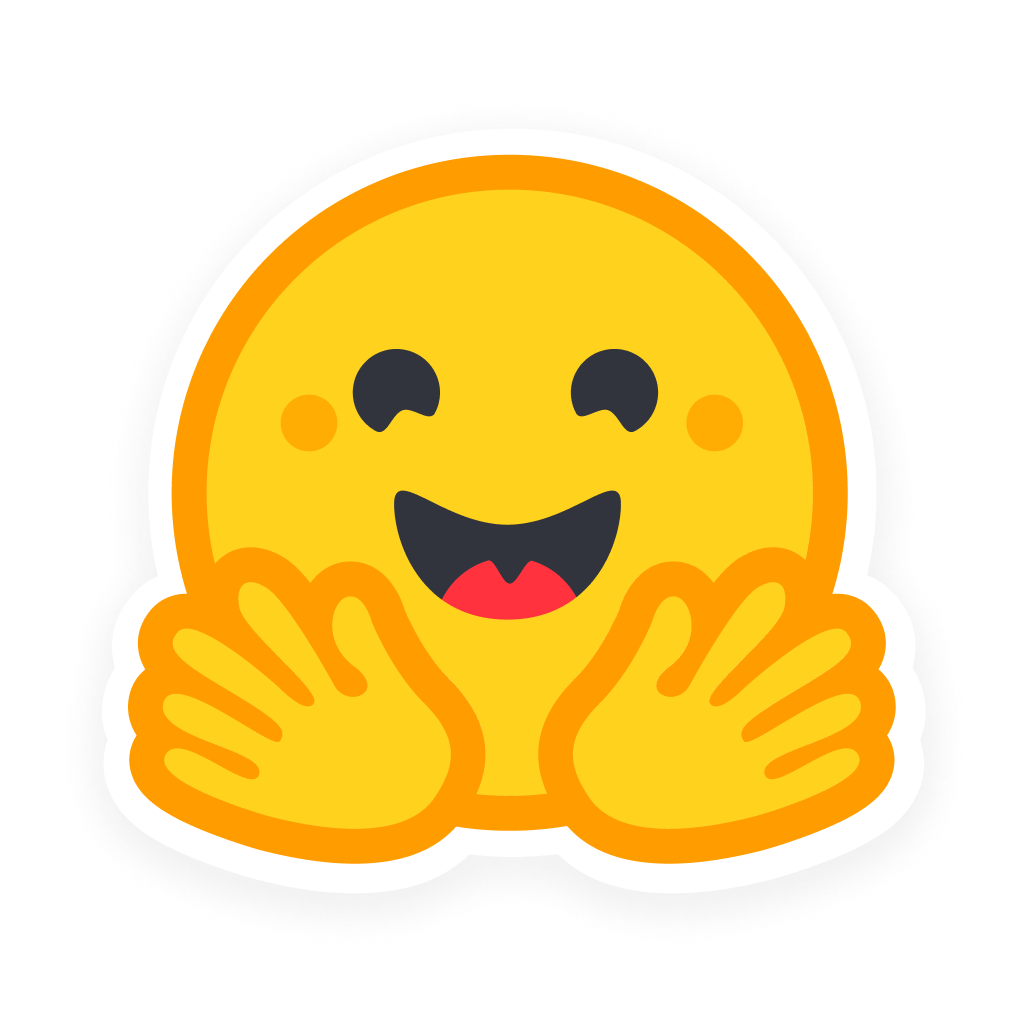} &
  \footnotesize{\texttt{\href{https://huggingface.co/collections/davidafrica/pico-maml-68e7703c4a40343e852db50f}{davidafrica/pico-maml}}} \\
  \cincludegraphics[width=1.5em]{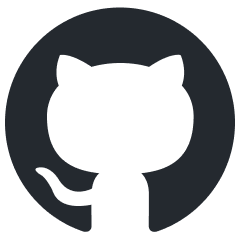} &
  
\footnotesize{\texttt{\href{https://github.com/DavidDemitriAfrica/pico-maml-train/tree/main}{DavidDemitriAfrica/pico-maml-train}}}
\end{tabular}

\end{abstract}

\section{Introduction}
Small language models (SLMs) are attractive for privacy and energy
reasons, but trail large models partly because they converge slowly and plateau early \citep{godey2024why, pmlr-v202-biderman23a, martinez2024}. As opposed to the common method of brute-force scaling, we explore a different axis: learning rules.  First-order Model-Agnostic Meta-Learning (MAML) \citep{pmlr-v70-finn17a} promises a learn-to-learn initialization, yet has rarely been applied to decoder models, and its effect on learning dynamics are poorly understood.

We address this by adding meta-learning in model pretraining\footnote{Using a lightweight modification of \textsc{pico-train} \citep{pico2025}, a language model pretraining suite.}, interleaving ordinary next-token loss (keeps fluency) with 32-way subset-mask \citep{bansal-etal-2020-self, li-zhang-2021-semi} episodes (forces rapid binding). Only a tiny MLP head is
adapted in the inner loop, so we can track backbone weights
without gradient noise. Our contributions are:

\begin{enumerate}
\item Four open SLMs (11M → 570M) trained with this \textcolor{mamlcolor}{\textbf{hybrid MAML rule}}.
\item A public trainer that logs per-checkpoint \textcolor{dynamicscolor}{\textbf{singular-value spectra}}, head entropies and query accuracy.
\item A candid evaluation on Universal NER: \textcolor{improvementcolor}{\textbf{modest gains at medium/large scales}} (+2–3 pp), alongside a \textcolor{degradationcolor}{\textbf{perplexity trade-off}}.
\item Observational evidence of a \textcolor{phasecolor}{\textbf{diversify–then–compress phase transition}} in effective rank.
\end{enumerate}

\paragraph{Reporting and scope.} All pretraining and fine-tuning results are from a single shared seed per condition due to compute limits; we therefore report averages across datasets where applicable, avoid statistical claims, and treat learning-dynamics findings as exploratory. We limit generalization claims to NER and to our training regime.

\section{Related Work}\label{sec:rw}
\textbf{Meta-learning for NLP.}
(MAML; \citealp{pmlr-v70-finn17a}) is an optimisation‑based form of meta‑learning that learns an initialisation from which a few gradient steps solve new tasks. It has been particularly successful in computer vision classification and reinforcement learning settings \citep{nichol2018first}. Within NLP, MAML has been adapted to a wide spectrum of supervised problems—including text classification, natural language inference, question answering, summarisation and named entity recognition---where a pre‑trained encoder such as BERT is further fine‑tuned on small datasets \citep{rajeswaran2019meta, raghu2021meta, hou-etal-2022-meta}. These studies operate (i) on encoder‑only, masked‑language models and (ii) at parameter counts close to the original 110M-parameter \textsc{BERT}. They leave open whether optimisation‑based meta-learning helps decoder LMs and whether its benefits persist at larger parameter scales.

\textbf{Meta-learning for pretraining.}
Initial NLP attempts applied MAML only at fine-tuning scale
\citep{raghu2021meta,hou-etal-2022-meta}.  More recent work embeds
bilevel objectives directly in pre-training
\citep{miranda2023pretrainingtrulybettermetalearning,ke-etal-2021-pre}. While promising, these efforts evaluate only a single model size, focus on one downstream task, or release neither code nor weights, limiting reproducibility and obscuring scale trends. We embed meta-learning directly into the pretraining loop, evaluate on various unseen domains in an unseen task, and provide open weights (11M-570M) and layer-wise spectra, filling that gap.

\textbf{Subset-Mask LMs (SMLMT).}
SMLMT constructs pseudo-tasks using a subset of vocabulary words \citep{bansal-etal-2020-self}.  Given an unlabeled text corpus, one selects a set of $N$ words and builds an $N$-way classification task. For each chosen word, sentences containing it are collected and the word is masked out. The task is then to predict the masked word from the $N$ candidates. \citet{li-zhang-2021-semi} interleaves it with ProtoNet tasks; we interleave with vanilla LM updates and scale to 570M params.

\textbf{Interpretable training dynamics.}
Various works discuss the training of language models in phase transitions \cite{olsson2022context, hoogland2024developmental}, describing broad changes in indicators as the model gains rapidly in capabilities over a short period of time. We study such phase transitions in the context of meta-learning in pretraining.

Effective-rank probes (entropy of singular values) highlight learning behavior in deep nets \citep{martinez2024}. Lower-rank structure and rank compression are well documented in the literature \citep{huh2021low, galanti2022sgd, jaderberg2014speeding}, and we focus on the timing and co-evolution of the measurements of effective-rank probes with episodic generalization under the hybrid objective (§\ref{sec:eval}).

\section{Method}\label{sec:method}

We pretrain four decoder models at 11M, 65M, 181M and 570M parameters with a hybrid objective \citep{li-zhang-2021-semi} that alternates conventional next-token prediction and first-order MAML episodes \citep{pmlr-v70-finn17a}. The episodes are generated with Subset-Masked Language Modelling Tasks (SMLMT) \citep{bansal-etal-2020-self}.   This section details the backbone, the meta-learning episode, the optimisation schedule, and the downstream evaluation harness.

% \subsection{Pico}
% Our experiments are built on top of the Pico language development framework. Pico is designed to support scientific develops of language models by intergrating a close-circuit training and analysis loop. ... 

\subsection{Baselines}
The starting point is the open Pico decoder \cite{pico2025}, a LLAMA-style \cite{touvron2023llama} stack implemented in plain PyTorch. To maintain apples-to-apples comparability with the original models (and as such isolate the effect of introducing MAML to pretraining), we maintain the design choices and hyperparameter choices of the original Pico decoder models. A sequence of $L=12$ decoder blocks receives 2048 input tokens.  Each block performs RMSNorm \cite{zhang2019rms}, grouped-query self-attention \cite{ainslie2023gqa} with rotary position embeddings \cite{su2024rope}, and a SwiGLU feed-forward network \cite{shazeer2020glu} that expands to $4d$ before projecting back to the model width $d$.  Width is the only scale-dependent hyper-parameter: $d\in\{96,384,768,1536\}$ for the tiny, small, medium and large variants.  All models use 12 heads, 4 key–value heads and causal masking.

\subsection{Task construction via SMLMT}\label{sec:smlmt}
SMLMT converts unlabelled text into few-shot classification tasks.  From the corpus we sample a set of $N$ content words, collect sentences that contain each word and replace that word with a single \texttt{<mask>}.  The goal is to predict which of the $N$ candidates was masked.  Each episode supplies $K$ support sentences and a disjoint query set.  Table~\ref{tab:smlmt-example-k2} shows an episode with $N=4$ city names and $K=2$ supports per class; the query asks the model to complete a new sentence about cherry blossoms.  In practice we use $N=32$ and $K=4$ so the task entropy matches the five-bit next-token uncertainty of English text\footnote{Shannon’s estimate of printed-English entropy is about 1.3 bits per character \citep{shannon1951prediction}; since English BPE tokens span on average about 4 characters \citep{openai_tokenization}, this implies roughly $\approx 5.2\ \text{bits/token}.$  We therefore use 5 bits per token as a conservative rule of thumb.}.

\begin{table}[t]
  \centering
  \small
    \resizebox{.5\textwidth}{!}{  
    \begin{tabular}{@{}llc@{}}
    \toprule
    \textbf{Set} & \textbf{Input (masked)}                                        & \textbf{Label} \\ 
    \midrule
    \textbf{Support} ($K{=}2$ each) 
      & I visited \_\_ last summer.                                   & Tokyo    \\
      & The sushi festival in \_\_ was unforgettable.                  & Tokyo    \\
      & The Big Ben is in \_\_.                                       & London   \\
      & I caught the tube at \_\_ yesterday.                          & London   \\
      & The Seine runs through \_\_.                                  & Paris    \\
      & She admired the art at the Louvre in \_\_.                    & Paris    \\
      & The Forbidden City is in \_\_.                                & Beijing  \\ 
      & I sampled Peking duck in \_\_.                                & Beijing  \\ 
    \midrule
    \textbf{Query}                    
      & I plan to travel to \_\_ to see the cherry blossoms.          & Tokyo    \\ 
    \bottomrule
  \end{tabular}}
  \caption{Example SMLMT episode with $N{=}4$ classes and $K{=}2$ support sentences per class.}
  \label{tab:smlmt-example-k2}
\end{table}

\subsection{Optimiser, data, and monitoring}
Training runs for 6000 outer updates on four A100 GPUs, with the original Pico-decoder models evaluated at the checkpoint after 6000 steps. Each GPU streams micro batches of 256 sequences from the 30\,percent English subset of Dolma \cite{soldaini2024dolma} that is already tokenised and chunked by Pico \cite{pico2025}. The outer optimiser is AdamW with peak learning rate $3\times10^{-4}$, 2500-step warm-up and cosine decay.  Micro batches of 256 sequences are accumulated eight times giving an
effective batch of 2048 (1024 for the 11M model).  Every 100 steps we
evaluate Paloma perplexity \citep{magnusson2024paloma} and log the
singular values of three attention and three feed-forward matrices to
compute effective rank \citep{martinez2024}.  Query and support
accuracies are also tracked.

\subsection{Downstream protocol}
Named entity recognition (NER), the downstream task for this study, is a fundamental NLP task that identifies and categorizes entities (e.g., persons, organizations, locations) within unstructured text \citep{chinchor1997muc}, and is used in healthcare \citep{kundeti2016clinical, polignano2021comparing, shafqat2022standard}, law \citep{leitner2019fine,au2022ner, naik2023legal}, business \citep{putthividhya2011bootstrapped, alvarado2015domain, zhao2021bert}, and knowledge graph systems \cite{al2020named}. Specifically, we evaluate our models on Universal NER benchmark \citep{mayhew-etal-2024-universal}. UNER v1 comprises three categories of NER evaluation data, each built on top of Universal Dependencies (UD) \citep{nivre-etal-2016-universal, nivre-etal-2020-universal} tokenization and annotations: publicly available in-language treebanks, parallel UD (PUD) evaluation, and other eval-only sets (Appendix \ref{app:uner}). 

After pretraining we load the checkpoint at step 6000 and attach a fresh linear classifier for UniversalNER. Two fine-tuning settings are used: head-only and full. In the head-only setting the Transformer is frozen so fine-tuning mirrors the inner loop, in the full setting all weights update.  Fine-tuning uses AdamW at $3\times10^{-5}$ for at most ten epochs with early stopping on development F$_1$.  

\section{Model Pretraining}
\begin{figure}[!t]
\centering
\includegraphics[width=1\linewidth]{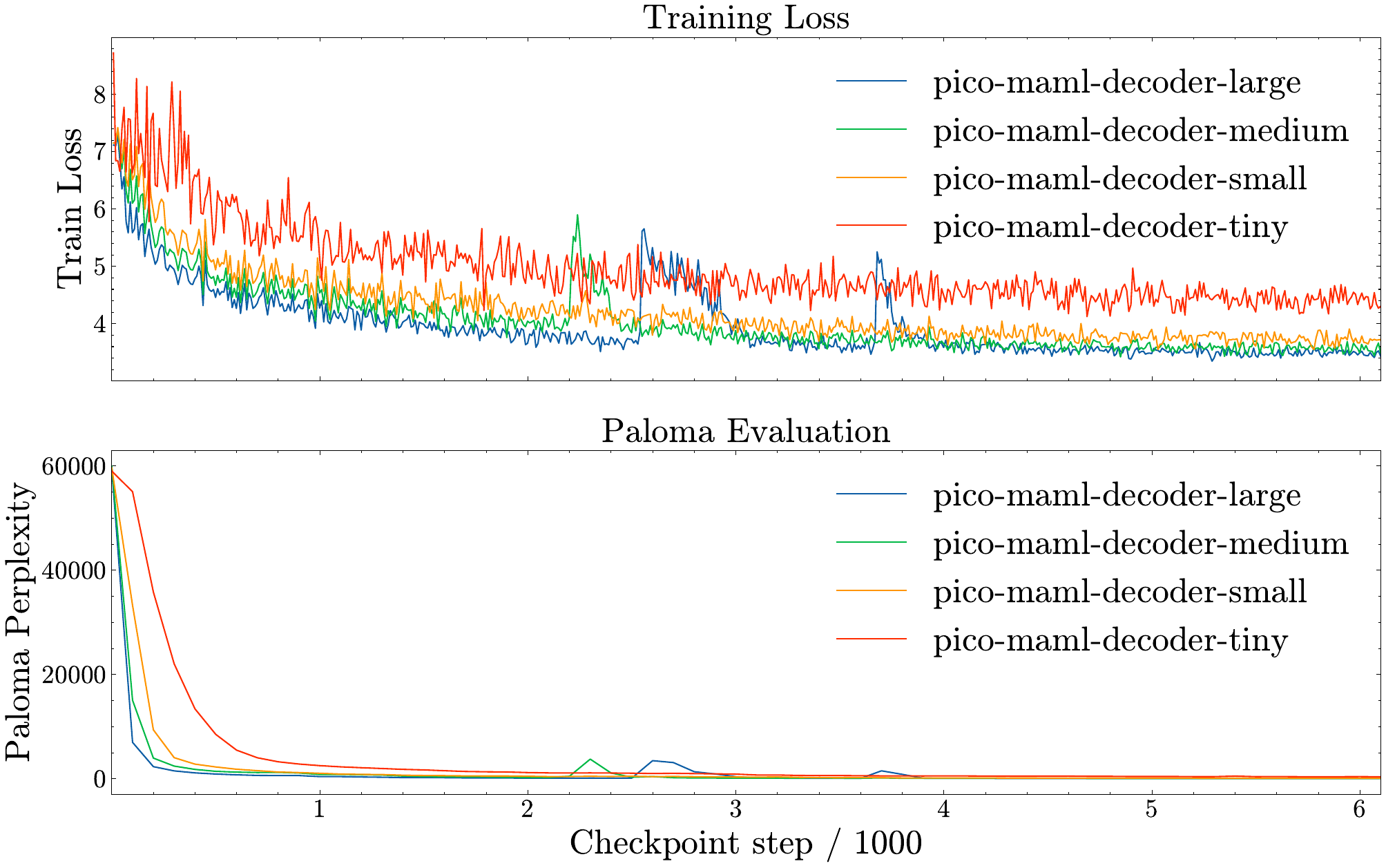}
\caption{Training loss and Paloma perplexity across pretraining steps for all MAML models. Two‐panel plot showing the evolution of (top) cross‐entropy training loss and (bottom) Paloma perplexity, each as a function of global pretraining step.} %Top (Training Loss): All models rapidly reduce loss from $\sim$8 to $\sim$4 within the first few hundred steps and then slowly converge toward $\sim$3.8 by step 6 000. Larger decoders attain lower final loss than smaller ones. Bottom (Paloma Perplexity): Perplexity plummets by $\sim 1000-2000$ steps for large and medium decoders. Ultimately, models stratify by capacity: larger decoders achieve the lowest perplexity under MAML meta-training.}
\label{fig:trainloss}
\end{figure}

\paragraph{Training-perplexity tradeoff across scales.} The prerequisite for modifying a pretraining method is ensuring the model still learns. All four Pico-MAML variants reach their respective vanilla loss 1.3–1.6× sooner (faster optimization), but Paloma perplexity is worse at most scales by 6000 steps (Table~\ref{tab:pretraining-metrics}).

\begin{table}[!t]
\centering
\small
\resizebox{0.47\textwidth}{!}{
\begin{tabular}{@{}lcc@{}}
\toprule
\textbf{Model} & \textbf{Train Loss @6k} & \textbf{Paloma Perplexity @6k} \\
\midrule
\texttt{pico-decoder-tiny} & 5.31 & 786.85 \\
\texttt{pico-maml-decoder-tiny} & \textbf{4.44} & \textbf{422.42} \\
\addlinespace
\texttt{pico-decoder-small} & 4.14 & \textbf{80.25} \\
\texttt{pico-maml-decoder-small} & \textbf{3.67} & 113.76 \\
\addlinespace
\texttt{pico-decoder-medium} & 3.89 & \textbf{77.90} \\
\texttt{pico-maml-decoder-medium} & \textbf{3.49} & 78.63 \\
\addlinespace
\texttt{pico-decoder-large} & 3.69 & \textbf{49.86} \\
\texttt{pico-maml-decoder-large} & \textbf{3.49} & 66.62 \\
\bottomrule
\end{tabular}}
\caption{For each model (rows) under vanilla vs. MAML pretraining (columns), shows cross-entropy loss and Paloma perplexity measured at exactly 6000 steps.} %Lower values indicate better convergence (loss) and language modeling quality (perplexity). Note that MAML variants achieve lower loss but mixed perplexity results, suggesting a trade-off between optimization speed and fluency.}
\label{tab:pretraining-metrics}
\end{table}

Contrary to expectation, MAML’s inductive bias may favor optimization over regularization. MAML accelerates convergence but degrades out-of-task fluency at most scales. However, this pattern is consistent with known multi-task interference: the episodic discriminative objective improves adaptation signals but can conflict with next-token distributional modeling under fixed compute and a single set of hyperparameters \cite{kendall2017multitask, yu2020gradientsurgerymultitasklearning, standley2020taskslearnedmultitasklearning}. Hence, it is unclear if the perplexity gap is an objective-mixing artifact or evidence that meta-learning inherently harms LM fluency.

\section{Downstream NER Evaluation}

% Explain the NER Task a bit 

\label{sec:eval}

Models are fine-tuned on each dataset in Universal NER \citep{mayhew-etal-2024-universal, nivre-etal-2016-universal, nivre-etal-2020-universal} with publicly available train and dev sets \footnote{Namely, \texttt{ddt, ewt, set, bosque, snk, set, talbanken, gsd, gsdsimp, all}.} Results (averaged across each finetuning dataset) are shown as micro-F1 scores in Table \ref{tab:summary_f1}, organized by evaluation group: seen (language with full train/test/dev splits), test-only (using Parallel Universal Dependencies PUD), and test-only low-resource languages (e.g., Cebuano, Tagalog). We report delta F1 as percentage points (pp) unless explicitly marked as percent change (\%).

\begin{table}[!t]
\centering
\resizebox{0.47\textwidth}{!}{
\begin{tabular}{lcccccc}
\toprule
\textbf{Model} 
  & \multicolumn{2}{c}{\textbf{Seen}} 
  & \multicolumn{2}{c}{\textbf{Test-Only (PUD)}} 
  & \multicolumn{2}{c}{\textbf{Test-Only (Other)}} \\
  & Head & Full & Head & Full & Head & Full \\
\midrule
\texttt{tiny} (\%)   
  & \cellcolor{red!20}-8.3  
  & \cellcolor{red!20}-3.0  
  & \cellcolor{green!20}+6.7  
  & \cellcolor{green!20}0.0   
  & \cellcolor{red!20}-37.5 
  & \cellcolor{green!20}+3.8  \\

\texttt{small} (\%)  
  & \cellcolor{green!20}+2.2  
  & \cellcolor{green!20}0.0   
  & \cellcolor{red!20}-17.2 
  & \cellcolor{red!20}-0.6  
  & \cellcolor{green!20}+46.7 
  & \cellcolor{green!20}+7.0  \\

\texttt{medium} (\%) 
  & \cellcolor{green!20}+1.9  
  & \cellcolor{green!20}+2.3  
  & \cellcolor{red!20}-4.6  
  & \cellcolor{green!20}+1.8  
  & \cellcolor{green!20}+14.8 
  & \cellcolor{green!20}+3.8  \\

\texttt{large} (\%)  
  & \cellcolor{green!20}+6.2  
  & \cellcolor{green!20}+4.8  
  & \cellcolor{green!20}+7.2  
  & \cellcolor{green!20}+3.5  
  & \cellcolor{green!20}+2.1  
  & \cellcolor{green!20}+8.1  \\
\bottomrule
\end{tabular}}
\caption{Relative percentage improvement of micro-F1 (higher = better) for head-only vs.\ full fine-tuning across seen, test-only (PUD), and low-resource language groups (other). Demonstrates MAML’s consistent 2–3 pp lift at medium/large scales under full tuning. \textcolor{improvementcolor}{\textbf{Green cells}} indicate MAML improvements; \textcolor{degradationcolor}{\textbf{red cells}} show degradations.}
\label{tab:summary_f1}
\end{table}

The most striking takeaway from this stage is that, when averaged across all evaluation steps in a category, absolute F1 remains low ($\leq 0.35$, i.e., $\leq 35\%$) due to poor zero-shot transfer, especially for logographic scripts. Overall, MAML improves mean uplift is approximately +2–3 pp when averaged over all in-language datasets at medium/large scales, confirming a modest “learning-to-learn” effect under full adaptation\footnote{While these results are much worse in comparison to the baseline in the original Universal NER paper \citep{mayhew-etal-2024-universal}, this is likely because $\text{XLM-R}_{\text{large}}$ is a multilingual model \citep{conneau-etal-2020-unsupervised} and the pretraining dataset for Pico is entirely in English.}. 

% \begin{figure}[!htbp]
% \centering
% \includegraphics[width=1\linewidth]{graphics/maml_large_heatmap_head.png}
% \caption{F1 scores on the Universal NER benchmark, grouped by evaluation category for pico-maml-decoder-large at 6k steps, fine-tuned in the full regime. Each cell reports the performance for the model fine-tuned on the row dataset and evaluated on the column dataset.}
% \label{fig:maml_large_head}
% \end{figure}

\paragraph{In-language NER gains suggest capacity-dependent meta-learning.} To better understand how meta-initialization influences cross-lingual transfer on seen languages, F1 scores are broken down by dataset within the in-language group. The results are separated by tuning regime to clarify the extent to which meta-learned representations help when only the classifier is updated (head-only) versus when the entire model is fine-tuned.

\label{sec:inlang_split}

In the head-only setting (Table \ref{tab:inlang_head}), absolute F1 scores remain low across most datasets. Tiny models fail to generalize altogether. MAML shows the strongest and most consistent gains at large scales (Table \ref{tab:inlang_head_relative_shortened})---most prominently on \texttt{en\_ewt}, \texttt{hr\_set}, and \texttt{sv\_talbanken}-suggesting that episodic pretraining creates more adaptable feature spaces, particularly for common entity types and scripts. On Chinese (\texttt{zh\_gsd}, \texttt{zh\_gsdsimp}), performance is uniformly poor, confirming the baseline result in \cite{mayhew-etal-2024-universal} that transfer from phonographic to logographic scripts is difficult.

\begin{table}[!t]
\centering
\resizebox{0.45\textwidth}{!}{%
\begin{tabular}{lrrrrr}
\toprule
\textbf{Model} 
  & \multicolumn{1}{c}{\textbf{Danish}} 
  & \multicolumn{1}{c}{\textbf{English}} 
  & \multicolumn{1}{c}{\textbf{Croatian}} 
  & \multicolumn{1}{c}{\textbf{Portuguese}} 
  & \multicolumn{1}{c}{\textbf{Swedish}} \\
\midrule
\texttt{large} (\%) 
  & \cellcolor{green!20}+8.1 
  & \cellcolor{green!20}+14.8 
  & \cellcolor{green!20}+10.7 
  & \cellcolor{green!20}+8.6  
  & \cellcolor{green!20}+18.0  \\
\bottomrule
\end{tabular}}
\caption{Percentage relative improvement of MAML over vanilla for head‐only tuning in the large model.}
\label{tab:inlang_head_relative_shortened}
\end{table}

\begin{table}[!t]
\centering
\resizebox{0.45\textwidth}{!}{%
\begin{tabular}{lccccc}
\toprule
Model & \textbf{Danish} & \textbf{English} & \textbf{Croatian} & \textbf{Portuguese} & \textbf{Swedish} \\
\midrule
\texttt{tiny} (\%)   
  & \cellcolor{green!20}+3.4  
  & \cellcolor{green!20}+0.2  
  & \cellcolor{red!20}-1.6  
  & \cellcolor{red!20}-0.7  
  & \cellcolor{green!20}+6.1  \\

\texttt{small} (\%) 
  & \cellcolor{red!20}-3.9  
  & \cellcolor{red!20}-4.7  
  & \cellcolor{red!20}-1.9  
  & \cellcolor{red!20}-2.6  
  & \cellcolor{green!20}+4.9  \\

\texttt{medium} (\%) 
  & \cellcolor{green!20}+0.8  
  & \cellcolor{green!20}+4.8  
  & \cellcolor{green!20}+3.9  
  & \cellcolor{green!20}+1.2  
  & \cellcolor{green!20}+3.7  \\

\texttt{large} (\%) 
  & \cellcolor{green!20}+3.6  
  & \cellcolor{green!20}+4.4  
  & \cellcolor{red!20}-0.5  
  & \cellcolor{green!20}+4.2  
  & \cellcolor{green!20}+2.8  \\
\bottomrule
\end{tabular}}
\caption{Percentage-wise relative improvement of MAML over vanilla under full tuning for each language.}
\label{tab:relative_improvement_inlang_shortened}
\end{table}

In the full setting (Table \ref{tab:relative_improvement_inlang_shortened}), both vanilla and MAML-pretrained models achieve higher F1 scores across the board. MAML confers consistent +0.01-0.03 gains at medium and large scales, especially for structurally complex languages like Croatian. These relative gains grow as model capacity increases, indicating that larger models benefit more from MAML pretraining. Even in Chinese, where scores are lowest, MAML nudges performance upward. These gains confirm that meta-pretraining does more than support shallow transfer: it reshapes the optimization landscape of the full model in a way that accelerates convergence and improves generalization. 

Taken together, these tables validate that MAML pretraining injects a scalable and tunable learning-to-learn signal. However, these average metrics do not tell the full story. Some settings, entity classes, and fine-tuning conditions benefit substantially more than others.

\begin{figure}[!t]
\centering
\includegraphics[width=1\linewidth]{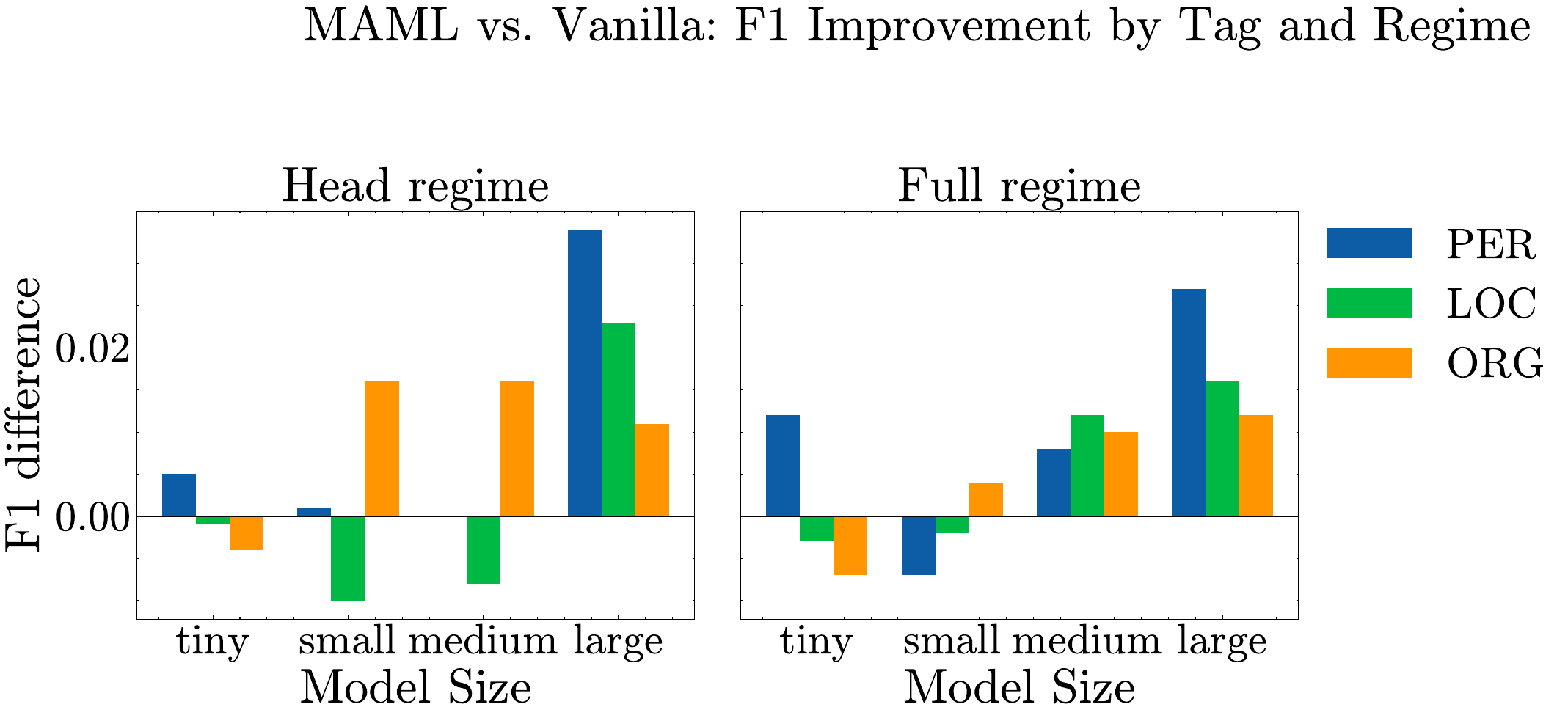}
\caption{MAML-Vanilla micro‐F1 difference by entity class and tuning regime, averaged across in-language datasets. Grouped bar charts reporting $\Delta$F1 = F1 MAML - F1 (Vanilla) for three named‐entity classes-PERSON (PER), LOCATION (LOC) and ORGANIZATION (ORG)-for pico-MAML decoders of four sizes (tiny, small, medium, large), averaged over nine in‐language NER datasets, over two fine-tuning regimes.}
\label{fig:f1_diff}
\end{figure}

\paragraph{Class-specific prototype bias in entity recognition.} We characterize the specific way MAML pretraining improves performance in NER by breaking down F1 score by entity class in Figure \ref{fig:f1_diff}.

Meta‐pretraining yields a clear capacity threshold in head‐only adaptation. Under a frozen backbone, only the large model consistently converts its learned initialization into PER (+0.034) and LOC (+0.023) gains; medium and smaller variants lack the representational bandwidth to rewire person and place distinctions via a shallow classifier. By contrast, even medium and small models see gains in ORG (+0.016 F1) likely because organization names often include distinctive tokens (e.g., “Inc.”, “Corp.”, or “University”) that form rigid, token‐level co‐occurrence patterns. These simple patterns mirror the pseudo-classification episodes SMLMT generates, so a shallow classifier can latch onto them without requiring deep feature reconfiguration.

Full fine‐tuning broadens and amplifies these effects. In the full setting, PER sees the largest MAML‐induced lift (up to +0.027 in the large model). LOC improvements (+0.016 at large scale) climb more gradually: place names often span heterogeneous contexts and scripts (e.g. Zagreb vs. Beijing), so meta‐pretraining must be supplemented by full gradient flow for location‐specific embeddings. ORG continues to enjoy gains (+0.012 at large), reinforcing that organization recognition remains the simplest class to bootstrap from episodic tasks.

\paragraph{Significant zero-shot transfer gains in low-resource languages.} Now, we discuss how inductive biases manifest in zero‐shot cross‐lingual transfer to low‐resource languages---namely, Tagalog (\texttt{tl}) and Cebuano (\texttt{ceb}).

Tagalog and Cebuano are the two most widely spoken native languages in the Philippines, with tens of millions of first-language speakers each. Both are typologically Austronesian and low-resource, but differ significantly. Tagalog is a morphologically rich, predicate-initial language with a complex voice system that encodes syntactic roles (agent, patient, locative, etc.) through verbal affixes and aspect-marking \citep{Kroeger-1993, Schachter-Otanes-1983, Ramos-2021-Tagalog}. Word order is flexible and often pragmatically driven, which weakens the utility of positional cues for tasks like named entity recognition. Cebuano is similarly Austronesian but morphologically simpler than Tagalog, with fewer voice alternations and less affixal variation \citep{Tanangkingsing-2011-Cebuano}. It also does not consistently mark syntactic roles with overt case particles; entities must be inferred from context rather than surface markers \citep{Sityar2000}. Additionally, Cebuano exhibits a distinct orthographic tradition and more conservative vocabulary (e.g., less Spanish borrowing) \citep{Bunye-Yap-1971-Cebuano}, which further distances it from the English-centric token distributions that dominate cross-lingual pretraining datasets. These characteristics make them ideal stress tests for testing the inductive bias of pretraining strategies like MAML.

\begin{table}[!t]
\centering
\small
\resizebox{0.5\textwidth}{!}{%
\begin{tabular}{llrrrr}
\toprule
\textbf{Model} & \textbf{Regime} & \textbf{Overall} & \textbf{Cebuano} & \textbf{Tagalog (TRG)} & \textbf{Tagalog (Ugnayan)} \\
\midrule
tiny   & head & \cellcolor{red!20}-100.0\% & \cellcolor{red!20}-100.0\% & N/A                 & N/A                  \\
small  & head & \cellcolor{green!20}+151.1\% & \cellcolor{green!20}+209.6\% & \cellcolor{green!20}+315.7\% & \cellcolor{red!20}-15.7\% \\
medium & head & \cellcolor{green!20}+24.3\%  & \cellcolor{green!20}+16.7\%  & \cellcolor{red!20}-20.7\%  & \cellcolor{green!20}+534.3\%\\
large  & head & \cellcolor{green!20}+9.0\%   & +0.0\%                     & \cellcolor{green!20}+57.3\%  & \cellcolor{red!20}-37.5\% \\
\addlinespace
tiny   & full & \cellcolor{red!20}-6.2\%    & \cellcolor{red!20}-4.7\%   & \cellcolor{red!20}-25.0\% & \cellcolor{green!20}+109.5\%\\
small  & full & \cellcolor{green!20}+7.3\%   & \cellcolor{red!20}-6.4\%   & \cellcolor{green!20}+28.8\%  & \cellcolor{green!20}+4.1\%  \\
medium & full & +0.0\%                     & \cellcolor{red!20}-1.0\%   & \cellcolor{green!20}+1.4\%   & \cellcolor{red!20}-2.1\%   \\
large  & full & \cellcolor{red!20}-8.0\%    & \cellcolor{red!20}-14.5\%  & \cellcolor{red!20}-1.6\%   & \cellcolor{red!20}-0.8\%   \\
\bottomrule
\end{tabular}}
\caption{Percentage change of MAML over vanilla zero‐shot NER transfer (from English) F1 on low‐resource languages (OTHER).}
\label{tab:transfer_other_pct}
\end{table}

In the head-only setting, MAML delivers its greatest impact on small and medium models. For example, the small head jumps from 0.088 to 0.221 overall---an absolute gain of 0.133 F1---and sees particularly large lifts in Cebuano (+0.153) and Tagalog-TRG (+0.262). The medium head also benefits substantially, improving from 0.259 to 0.322. Even the large head picks up a modest +0.030 F1. Only the tiny head collapses, reflecting its inability to form reliable prototypes during meta-training. These patterns suggest that MAML’s episodic learning instills useful, language-agnostic representations in the classifier layers, enabling mid-size heads to generalize token-level cues to new languages without modifying the backbone.

Once we allow full fine-tuning, however, most of MAML’s advantages disappear at higher capacities. The small model retains a small +0.026 F1 edge, but the medium shows no net change and the large actually drops by 0.034. This reversal implies that when every parameter is free to update, the strong gradient signals of full fine-tuning quickly override the meta-learned inductive biases, erasing or even inverting MAML’s earlier head-only gains. The tiny model again underperforms, consistent with its tendency to overfit during meta-training when unconstrained by a fixed backbone.

In the UNER benchmark, Tagalog and Cebuano serve as canonical low-resource, typologically distinct evaluation settings. Overall NER performance remains modest, but, as Table~\ref{tab:transfer_other_pct} shows, MAML provides meaningful zero-shot boosts in the head-only regime for small and medium models. These gains suggest that even without training exposure to these languages, the inductive biases from English episodic training transfer surprisingly well, at least for token-level prototypes.

\section{Learning Dynamics}

Despite clear convergence gains, the pretraining metrics alone leave several observations unexplained: the mid-training rebound and double-descent in Paloma perplexity, the abrupt jumps in support versus query accuracy, and the sudden collapse in representation rank. To understand this further, we now turn to a learning-dynamics analysis: tracking episodic support/query performance, classifier head statistics, and proportional effective rank throughout pretraining. 

\begin{figure}[!t]
\centering
\includegraphics[width=1\linewidth]{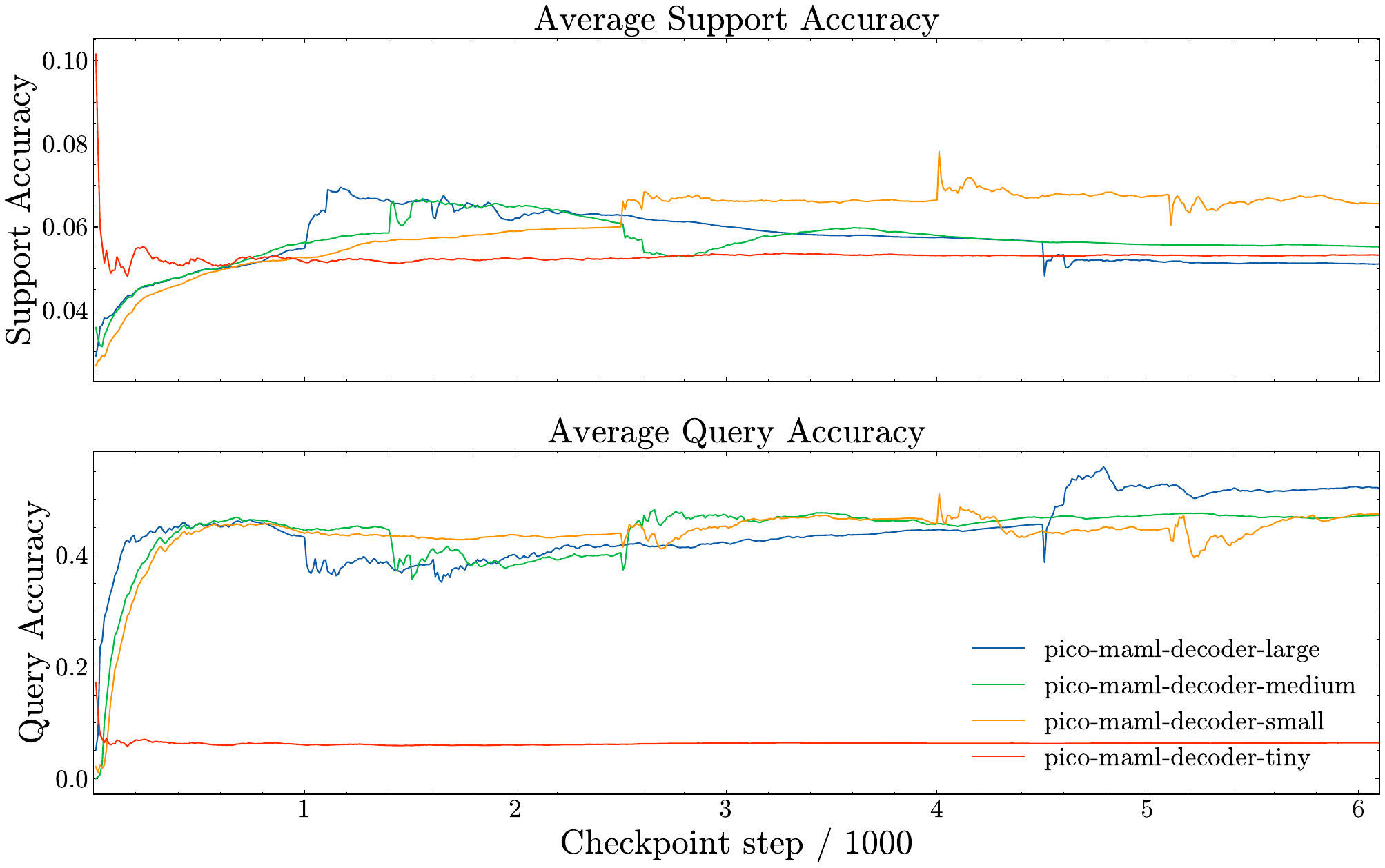}
\caption{Average support and query accuracy across pretraining steps for all models. Top: Average support‐set accuracy (\%) measured at the end of each inner‐loop adaptation, as a function of the global pretraining step. Bottom: Corresponding average query‐set accuracy (\%) after adaptation.} % The small and medium decoders exhibit classic meta‐learning behaviour, with support accuracy rising to 6-7 \% and query accuracy climbing steadily to $\geq40$ \%. The tiny decoder shows modest support improvements but stagnating query performance, indicative of insufficient representational capacity. The large decoder attains the highest absolute accuracies ($\geq50$ \%) but with pronounced fluctuations in both support and query curves, reflecting greater optimizer noise at larger scale.}
\label{fig:meta}
\end{figure}

\paragraph{Effective meta-learning has a capacity threshold.} To understand how MAML updates influence learning dynamics during pretraining, we track both support (training set in the inner loop) and query (held out final step in the inner loop) accuracy across training steps (Figure~\ref{fig:meta}).

The \texttt{small} and \texttt{medium} models show clear signs of effective meta-learning. Support accuracy gradually increases and stabilizes around 6--7\%, while query accuracy climbs steadily above 40\%. This pattern indicates that the models are internalizing a useful task prior, and show smooth convergence with relatively little instability.

The \texttt{tiny} model displays a distinct failure mode. While its support accuracy rises modestly, its query accuracy remains stagnant, hovering just above chance (10\%). This suggests the model memorizes support examples but fails to learn task-generalizable features-a canonical symptom of underparameterization in meta-learning \citep{pmlr-v70-finn17a, rajeswaran2019meta}. In effect, it lacks the representational bandwidth to encode a shared inductive bias across tasks.

The \texttt{large} model shows a late-phase rise in query accuracy after 4,500 steps, coinciding with stabilization of head-weight variance. This suggests a phase-like reorganization where the model consolidates a useful episodic prior after a prolonged plateau. In the MAML setting, this may correspond to the model first learning how to adapt, before learning to generalize from adaptation.

Taken together, these patterns confirm that meta-learning is most stable within a mid-capacity regime. Models must be large enough to encode reusable structure, but not so large that their learning becomes erratic. These insights help contextualize downstream findings: the best generalization often arises from models that strike a balance between representational power and stable task-level adaptation.

\begin{figure}[!t]
\centering
\includegraphics[width=1\linewidth]{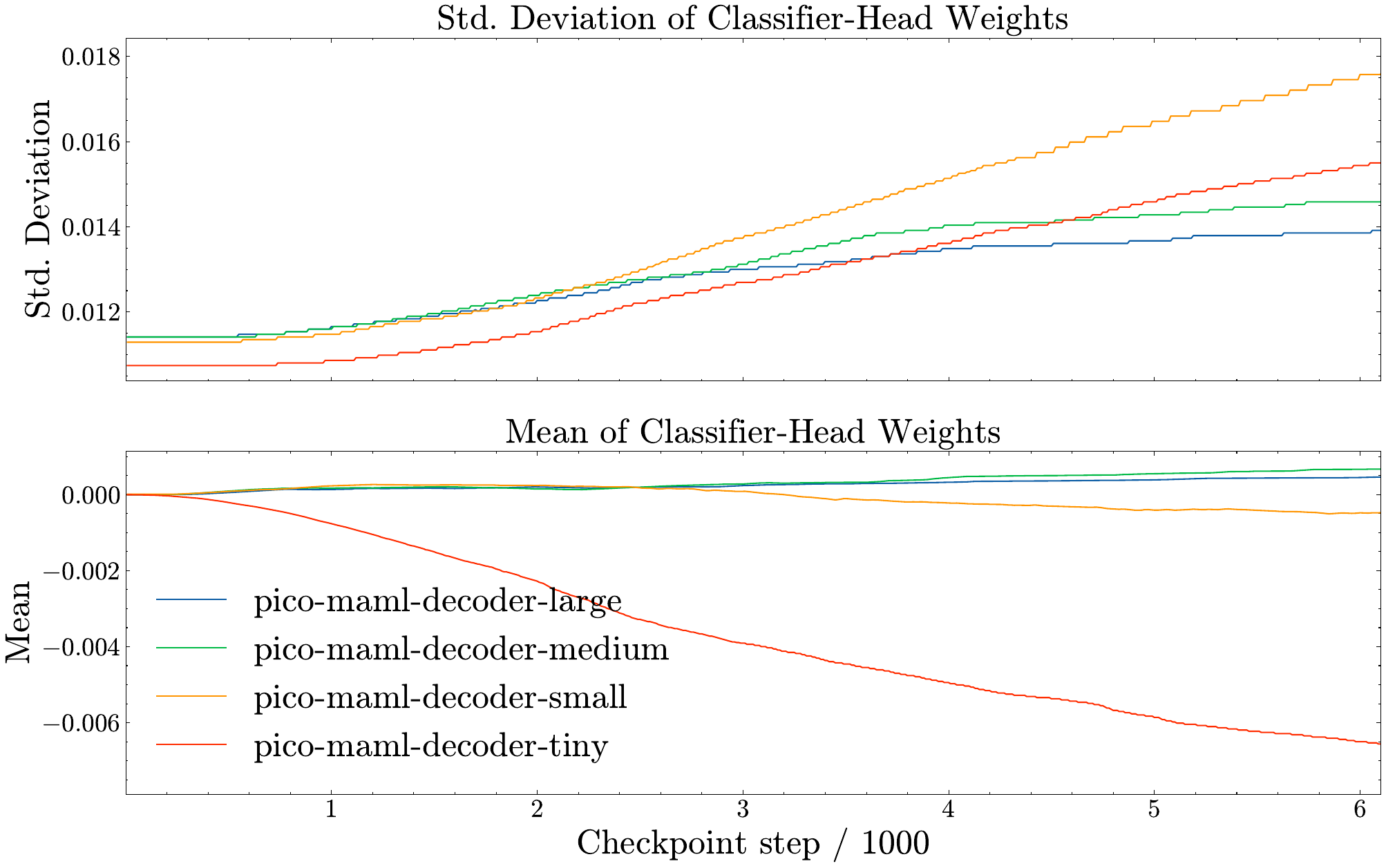}
\caption{Evolution of classifier head weights during meta-training. Top: Standard deviation of the task-specific classifier head weights (in logits space). Bottom: Mean of the classifier head weights.} % Although the head is re-initialized before each episode, its accumulated long-term statistics reveal how much the outer loop consolidates across tasks. All models increase their weight variance over time-especially the small decoder, whose variance surpasses all others after $\sim$2k steps, indicating an over-specialization tendency-while the tiny decoder develops a strong bias (nonzero mean) without commensurate variance, reflecting representational collapse.}
\label{fig:head}
\end{figure}

\paragraph{Classifier head weight variance reveals adaptation behavior.} To probe how episodic adaptation reshapes the backbone’s feature geometry, we track the mean and standard deviation of the episodically adapted classifier head across training (Figure~\ref{fig:head}). Because the inner loop updates only this shallow head on frozen backbone features, its across-episode weight statistics act as a lightweight linear-probe proxy for class separability: under softmax on fixed features, class weight vectors tend to align with differences between class means, so greater dispersion (std) across head weights indicates larger between-class margins induced by the backbone, while transient spikes without sustained query gains suggest support overfit rather than stable generalization. We therefore relate inflections in mean/std to simultaneous changes in support/query accuracy to contextualize adaptation quality.

The top panel shows the standard deviation of head weights. All models exhibit growth in weight variance, indicating increasing expressivity in the task-specific head. The \texttt{small} model diverges most sharply, with its weight variance surpassing all others after 2k steps. This suggests an over-specialization effect: the model learns to adapt aggressively to each task, potentially at the cost of stability. 
In the lower panel, the mean of the head weights remains near zero for most models, but the \texttt{tiny} model is an outlier. It accumulates a strong bias in one direction over training, indicating that its head converges toward a fixed mapping that is minimally updated across episodes. This aligns with earlier diagnostics showing that its gradient norms collapse early in training.

These dynamics reinforce the idea that episodic MAML indeed induces a scale-sensitive tradeoff: in higher-capacity models, episodic gradients drive generalizable structure into the shared initialization; in lower-capacity models, this same pressure can cause drift or collapse.

\begin{figure}[!t]
\centering
\includegraphics[width=1\linewidth]{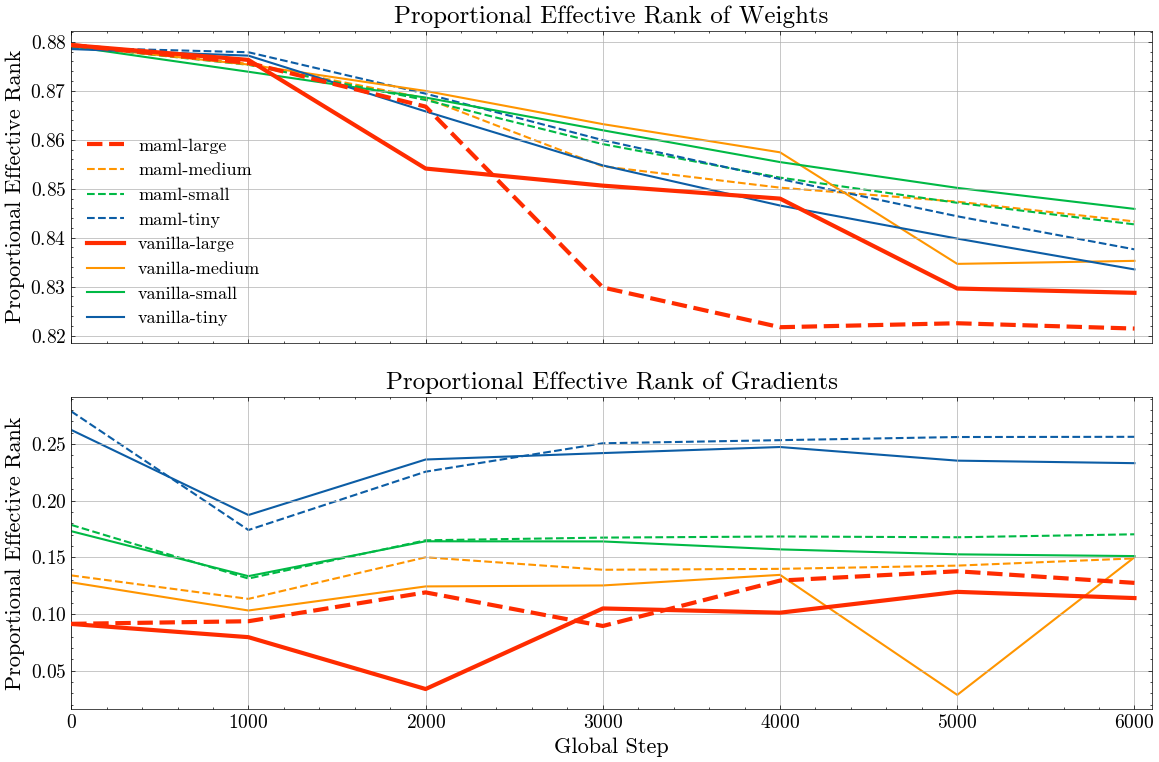}
\caption{Proportional effective rank of MAML and vanilla models on available checkpoints until 6k steps. Top: weights; bottom: gradients.}
\label{fig:per-er}
\end{figure}

\paragraph{Evidence of representation collapse and reorganization.} To understand how MAML alters internal representations, we track \textit{proportional effective rank} (PER), a structure-sensitive metric during training applied to both weights and gradients in the attention layers (Figure~\ref{fig:per-er}).

Following \citet{roy2007effective} and \citet{martinez2024}, effective rank measures the entropy of the singular value spectrum of a matrix, while PER normalizes this by the total dimensionality: 
$$\mathrm{PER}(W) = \frac{\exp\left(-\sum_i p_i \log p_i\right)}{d}$$ where $p_i = \frac{\sigma_i}{\sum_j \sigma_j}$. PER captures the extent to which the model’s representations or updates span a full-dimensional space; a decline in PER indicates compression or structural specialization.

\begin{tcolorbox}[colback=lightphase,colframe=phasecolor,boxrule=1pt,arc=2mm,title={\textcolor{white}{\textbf{Key Finding: Phase Transition in Large Model}}}]
Across all MAML-pretrained models, PER declines over training, but the \texttt{large} model exhibits an \textcolor{phasecolor}{\textbf{abrupt, synchronized drop}} at step ${\sim}3000$ in:
\begin{itemize}[nosep,leftmargin=*]
    \item \textcolor{phasecolor}{Proportional effective rank} (PER)
    \item \textcolor{vanillacolor}{Paloma perplexity} (after initial rise)
    \item \textcolor{mamlcolor}{Query accuracy} (sharp jump from plateau)
\end{itemize}
\end{tcolorbox}

We interpret this behavior as a \textcolor{phasecolor}{\textbf{representational phase transition}}: the model initially fits the objective using diffuse, high-dimensional representations, which are later compressed into task-specialized, low-rank structures. The descent in PER lags behind the initial perplexity gains, and only after this drop does the second descent in Paloma begin. There is no strong evidence of a comparable phase transition in the vanilla models. While the \texttt{large} and \texttt{medium} variants show mild inflection points in loss and perplexity around step 3000, these are gradual and lack the coordinated sharpness seen in the MAML-trained models. 

This suggests that MAML’s bilevel updates and episodic task pressure may help reorganize the optimization landscape to favor discrete qualitative shifts in representation. As explored in \citet{olsson2022context, wang2024loss, hoogland2024developmental}, model training often proceeds in qualitatively distinct stages: from brute-force fitting, to intermediate rule memorization, to compressed algorithmic abstraction. The drop in PER may signal such a transition---from early diffuse representations to compressed heads tuned to solve the repeated structure of SMLMT episodes. This representational transition is also reflected in the model’s adaptation performance. Around the same step where PER and Paloma perplexity undergo a sharp drop (step $\sim$3000), both support and query accuracies rise abruptly (see Figure~\ref{fig:meta}). Prior to this point, query accuracy remains relatively flat, indicating that the model struggles to generalize from support to query examples. But after the phase transition, the model rapidly learns to extrapolate, with query accuracy climbing from near random to over 0.5.

\begin{figure}[!tb]
    \centering
\includegraphics[width=\linewidth]{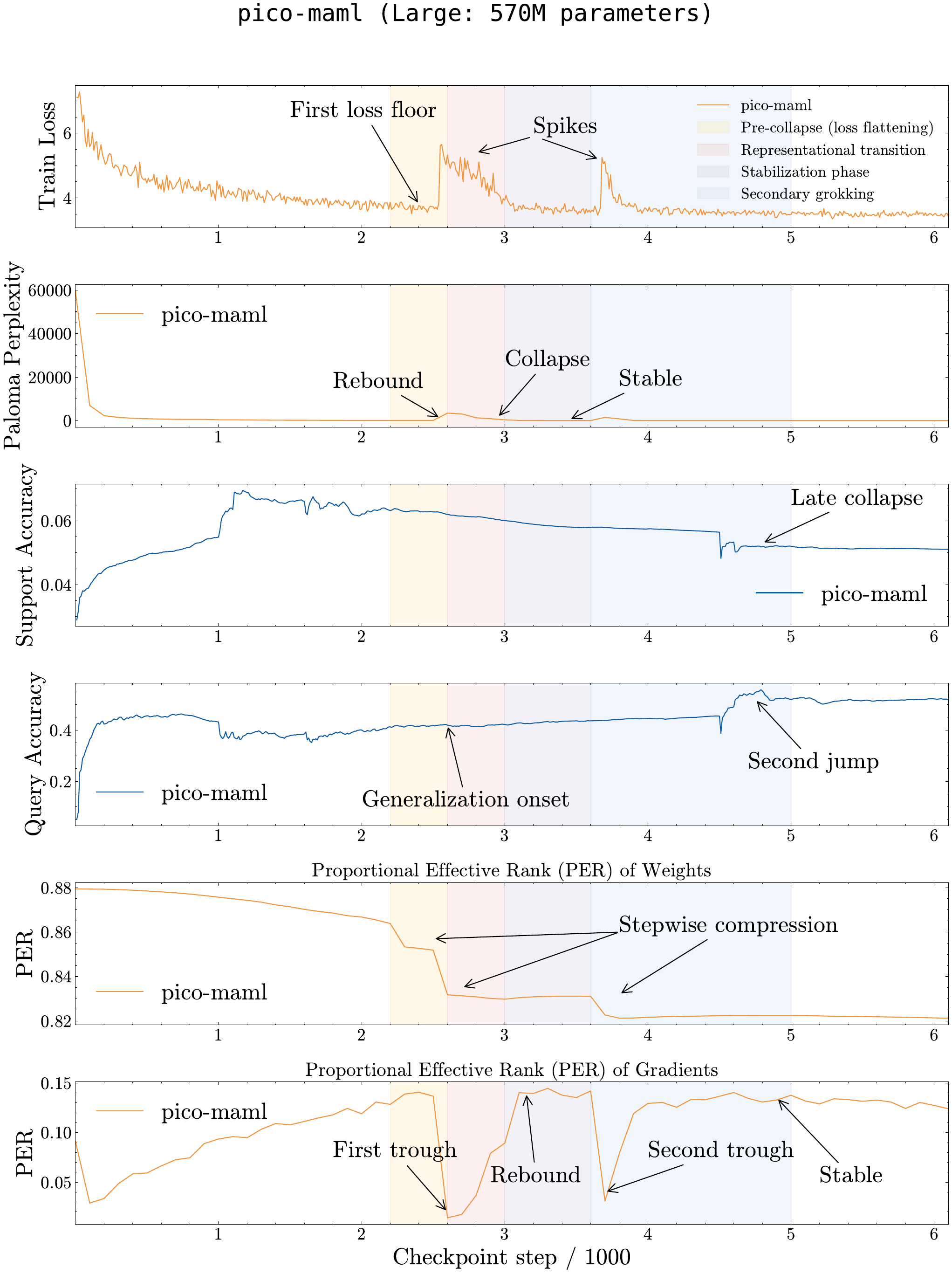}
\caption{Dynamics of \texttt{pico-maml-decoder-large} over 6000 pretraining steps. \textcolor{phasecolor}{\textbf{Pink shaded region}} marks the phase transition (steps 2600-3200) where PER collapses, perplexity drops, and query accuracy jumps.}
%(Top) Training loss: early plateau, phase-transition spike (2.8k), second spike (3.8k), and convergence. 
%(Second) Paloma perplexity: rebound, collapse (double descent), plateau, and second descent. 
%(Support \& query) Support accuracy plateaus then collapses; query accuracy generalizes at 2.6k with a second jump at 4.9k. 
%(Bottom) PER of weights and gradients: weights collapse without recovery; gradients show several troughs and rebounds. Shaded bands mark distinct regions.}
\label{fig:big-bosh}
\end{figure}

This synchrony across metrics provides compelling evidence of a coordinated phase shift in the model’s learning trajectory. When looking into more granular checkpoints (Figure \ref{fig:big-bosh}), there is clearer evidence that the model transitions from an early stage where it relies on diffuse representations to a later stage where it reorganizes both its representations and update paths into a lower-dimensional, more modular form capable of few-shot generalization. That said, this phase behavior appears scale-sensitive as it is absent in smaller scales. This suggests that the capacity to reorganize may be gated by scale, and that below a certain threshold, the inductive pressure of MAML induces collapse rather than modularization. 

\section{Conclusion}

We interleaved first–order MAML episodes with decoder pretraining and analyzed dynamics across four SLM scales. Under equal compute, the hybrid objective accelerates optimization but trades off perplexity at most scales; downstream, it brings modest average NER gains (+2–3 pp) at medium/large scales. Spectral logs expose a phase-like diversify–then–compress pattern that coincides with improving episodic query accuracy in the large model. Given our NER-only, single-seed scope, we present these as tools and observations rather than broad performance claims.

However, while our evaluation focuses exclusively on named entity recognition, the underlying mechanism---episodic adaptation via SMLMT---is task-agnostic. In principle, the same hybrid objective could be applied to other sequence labeling tasks (e.g., part-of-speech tagging, syntactic chunking) or even structured prediction problems that admit few-shot formulations. Whether the phase transitions and rank-compression patterns we observe generalize to non-linguistic domains (e.g., code generation, mathematical reasoning) remains an open question. Future work should explore whether MAML's inductive bias is inherently suited to token-level structure learning or whether it confers broader benefits across modalities and task families.

Relatedly, other natural extensions suggest themselves. Future work should also include multi-seed and hyperparameter sweeps (inner LR, episode frequency), multilingual pretraining to test cross-script transfer, varying which layers adapt in the inner loop, and evaluation on non-NER tasks (e.g., classification, QA, reasoning), as the architectural design space is rather large. In terms of exploratory work, a natural next step is to learn whether the same phase transition re-emerges when the corpus is multilingual, which would clarify why cross-script transfer remains the weak point of the present models. Varying which backbone layers adapt, how many steps they receive and how frequently episodes are interleaved may unlock better compute–capability trade-offs. Finally, the clear correlation between the effective-rank collapse and downstream utility hints that spectral diagnostics might serve as a self-supervised early-stopping signal.

\newpage

\section*{Limitations}

All training runs stop at exactly six thousand outer steps, a horizon that may be too short for the largest model, so the observed perplexity gap between MAML and vanilla training could shrink or even reverse if optimisation were allowed to continue. Our downstream evaluation focuses on a single task family, sequence labelling, so it remains unclear whether the same advantages would materialise on reasoning or generation-quality benchmarks. Because the corpus is predominantly English, improvements in low-resource or logographic languages remain modest; a more diverse corpus may alter both quantitative and qualitative conclusions. Hyper-parameters such as the hybrid episode probability, the inner-loop learning rate and the 32-way 4-shot episode size were transferred unchanged across scales; dedicated tuning might further modify the trade-off between convergence speed and final perplexity. Models were trained on academic budget, which limited training to 6000 outer steps. Some interesting training dynamics only appear after a very extended period of training, and future work should study this long-term behavior. Finally, each condition was run with a single random seed owing to GPU constraints, so although the phase transition appears robust, the exact magnitude of the gains should be interpreted with caution. 

\section*{Acknowledgments}

This work was supported by a grant from the Accelerate Programme for Scientific Discovery, made possible by a donation from Schmidt Futures. David Demitri Africa is supported by the Cambridge Trust and the Jardine Foundation. Richard Diehl Martinez is supported by the Gates Cambridge Trust (grant OPP1144 from the Bill \& Melinda Gates Foundation). Suchir Salhan is supported by Cambridge University Press \& Assessment.
\bibliography{custom}

\clearpage
\appendix

\section{Pseudocode}

Below is the pseudocode for the MAML and vanilla pretraining setup.

\begin{algorithm}\small \caption{Distributed SMLMT Loop}\label{alg:dist-maml} \begin{algorithmic} \State Initialize model $f_\theta$, head $h_\phi$, outer optimizer, and inner SGD on $h_\phi$ \State step $\leftarrow 0$ \For{each sub-batch $B$ from dataloader} \State $X \leftarrow \text{AllGather}(B\ \text{inputs})$ \Comment{across devices} \State $r \leftarrow \text{Broadcast}(\text{Uniform}(0,1))$ \If{$r < \rho$} \State $(S,Q,y_S,y_Q) \leftarrow \text{MaskTokens}(X)$ \State $\phi_{\text{snap}} \leftarrow \phi$ \Comment{save head params} \For{$t = 1$ to $T_{\text{inner}}$} \State $\ell_S \leftarrow \text{CE}\big(h_\phi(f_\theta(S)),,y_S\big)$ \State $\phi \leftarrow \phi - \alpha,\nabla_\phi \ell_S$ \EndFor \State $\ell \leftarrow \text{CE}\big(h_\phi(f_\theta(Q)),,y_Q\big)$ \State $\phi \leftarrow \phi_{\text{snap}}$ \Comment{restore head} \Else \State $X_{\text{in}} \leftarrow X$ without last token; $Y \leftarrow X$ without first token \State $\ell \leftarrow \text{CE}\big(f_\theta(X_{\text{in}}),,Y\big)$ \EndIf \State Backward$\big(\ell ,/, \text{accum\_steps}\big)$ \If{$( \text{step} + 1 ) \bmod \text{accum\_steps} = 0$} \State OptimizerStep(); SchedulerStep(); ZeroGrad() \State AggregateMetrics($\ell$); Barrier() \EndIf \State step $\leftarrow$ step $+ 1$ \EndFor \end{algorithmic} \end{algorithm}

\begin{algorithm}\small \caption{Distributed AR Loop}\label{alg:dist-ar} \begin{algorithmic} \State Initialize configs, Fabric/strategy, tokenizer, model $f_\theta$, optimizer \State Prepare dataloader and distribute it \State step $\leftarrow 0$; ZeroGrad() \For{each sub-batch $B$ from dataloader} \State $X \leftarrow \text{AllGather}(B\ \text{inputs})$ \Comment{across devices} \State $X_{\text{in}} \leftarrow X$ without last token; $Y \leftarrow X$ without first token \State $\ell \leftarrow \text{CE}\big(f_\theta(X_{\text{in}}),,Y\big)$ \State Backward$\big(\ell ,/, \text{accum\_steps}\big)$ \If{$( \text{step} + 1 ) \bmod \text{accum\_steps} = 0$} \State OptimizerStep(); SchedulerStep(); ZeroGrad() \State Barrier() \Comment{optional} \EndIf \State step $\leftarrow$ step $+ 1$ \EndFor \end{algorithmic} \end{algorithm}

\subsection{Multi-GPU processing}
Pico already uses Lightning-Fabric data parallelism but meta-learning introduces various demands that make multi-GPU processing complicated. A Bernoulli draw is done on one GPU and broadcast so all ranks choose the same objective. Support and query tensors are constructed on rank 0 then scattered, because per-rank random masks would destroy gradient equivalence. Every GPU performs the same ten head updates before any gradient is communicated.  A stray early \texttt{all\_reduce} would mix gradients from different inner steps, so we place an explicit \texttt{barrier} between inner and outer phases.

\section{Universal NER Datasets}
\label{app:uner}

To comprehensively evaluate the pretraining method, each permutation of fine-tuning setup (\{head-only, full\}, fine-tuning dataset (\{\texttt{da\_ddt}, \dots, \texttt{zh\_gsdsimp}, \texttt{all}\}) (where \texttt{all} consists of all available training sets), model size (\{tiny, small, medium, large\}), and pretraining setup (\{vanilla, MAML\}) is evaluated, for a total of 160 evaluation runs.

\begin{itemize}
  \item \textbf{Publicly Available In-language treebanks} (9 langs): full \texttt{train}/\texttt{dev}/\texttt{test} splits, identical to the official UD partitions.  
    \begin{itemize}
      \item \texttt{da\_ddt}, \texttt{en\_ewt}, \texttt{hr\_set}, \texttt{pt\_bosque}, \texttt{sk\_snk}, \texttt{sr\_set}, \texttt{sv\_talbanken}, \texttt{zh\_gsd}, \texttt{zh\_gsdsimp}
    \end{itemize}
  \item \textbf{Parallel UD (PUD) evaluation} (6 langs): single \texttt{test.txt} files, all sentence-aligned across German, English, Portuguese, Russian, Swedish and Chinese.  
    \begin{itemize}
      \item \texttt{de\_pud}, \texttt{en\_pud}, \texttt{pt\_pud}, \texttt{ru\_pud}, \texttt{sv\_pud}, \texttt{zh\_pud}
    \end{itemize}
  \item \textbf{Other eval-only sets} (3 langs): small test splits for low-resource languages.  
    \begin{itemize}
      \item \texttt{ceb\_gja} (Cebuano), \texttt{tl\_trg} (Tagalog TRG), \texttt{tl\_ugnayan} (Tagalog Ugnayan)
    \end{itemize}
\end{itemize}

\section{Supplementary Figures}

\subsection{Supplementary Tables}

\begin{table}[!htbp]
\centering
\caption{Micro-F1 scores (rows: selected datasets, columns: vanilla vs. MAML) under head-only tuning for large models. Highlights which languages benefit most from MAML without full adaptation.}
\label{tab:inlang_head}
\resizebox{0.5\textwidth}{!}{
\begin{tabular}{lccccccccc}
\toprule
Model 
  & da\_ddt & en\_ewt & hr\_set & pt\_bosque 
  & sk\_snk & sr\_set & sv\_talbanken 
  & zh\_gsd & zh\_gsdsimp \\
\midrule
vanilla\_tiny   & \textbf{0.004}& 0.031 & \textbf{0.011} & 0.000 & 0.004 & \textbf{0.009} & \textbf{0.000} & \textbf{0.005} & \textbf{0.009} \\
maml\_tiny      & 0.000 & \textbf{0.057} & 0.000 & \textbf{0.014} & \textbf{0.014} & 0.002 & \textbf{0.000} & 0.000 & 0.005 \\
\addlinespace
vanilla\_small  & 0.000 & \textbf{0.196} & 0.123 & 0.099 & 0.047 & \textbf{0.056} & \textbf{0.020} & 0.000 & 0.003 \\
maml\_small     & \textbf{0.004} & 0.156 & \textbf{0.162} & \textbf{0.104} & \textbf{0.063} & 0.044 & 0.000 & \textbf{0.003} & \textbf{0.005} \\
\addlinespace
vanilla\_medium & \textbf{0.141} & 0.252 & 0.311 & 0.240 & \textbf{0.153} & 0.325 & 0.065 & \textbf{0.010} & \textbf{0.020} \\
maml\_medium    & 0.087 & \textbf{0.288} & \textbf{0.329} & \textbf{0.243} & 0.136 & \textbf{0.362} & \textbf{0.108} & 0.005 & 0.010 \\
\addlinespace
vanilla\_large  & 0.247 & 0.366 & 0.401 & 0.337 & 0.178 & 0.422 & 0.261 & \textbf{0.034} & 0.039 \\
maml\_large     & \textbf{0.267} & \textbf{0.420} & \textbf{0.444} & \textbf{0.366} & \textbf{0.191} & \textbf{0.455} & \textbf{0.308} & 0.023 & \textbf{0.040} \\
\bottomrule
\end{tabular}
}
\end{table}

\begin{table}[!ht]
\centering
\caption{Percentage relative improvement of MAML over vanilla for head‐only tuning in the large model.}
\label{tab:inlang_head_relative}
\resizebox{0.5\textwidth}{!}{%
\begin{tabular}{lrrrrrrrrr}
\toprule
\textbf{Model} 
  & \multicolumn{1}{c}{da\_ddt} 
  & \multicolumn{1}{c}{en\_ewt} 
  & \multicolumn{1}{c}{hr\_set} 
  & \multicolumn{1}{c}{pt\_bosque} 
  & \multicolumn{1}{c}{sk\_snk} 
  & \multicolumn{1}{c}{sr\_set} 
  & \multicolumn{1}{c}{sv\_talbanken} 
  & \multicolumn{1}{c}{zh\_gsd} 
  & \multicolumn{1}{c}{zh\_gsdsimp} \\
\midrule
\texttt{Large} (\%) 
  & \cellcolor{green!20}+8.1 
  & \cellcolor{green!20}+14.8 
  & \cellcolor{green!20}+10.7 
  & \cellcolor{green!20}+8.6  
  & \cellcolor{green!20}+7.3  
  & \cellcolor{green!20}+7.8  
  & \cellcolor{green!20}+18.0  
  & \cellcolor{red!20}-32.4 
  & \cellcolor{green!20}+2.6  \\
\bottomrule
\end{tabular}}
\end{table}

\begin{table}[!htbp]
\centering
\caption{Percentage‐wise relative improvement of MAML over vanilla under full tuning for each language.}
\label{tab:relative_improvement_inlang}
\resizebox{0.5\textwidth}{!}{%
\begin{tabular}{lccccccccc}
\toprule
Model & da\_ddt & en\_ewt & hr\_set & pt\_bosque & sk\_snk & sr\_set & sv\_talbanken & zh\_gsd & zh\_gsdsimp \\
\midrule
\texttt{tiny} (\%)   
  & \cellcolor{green!20}+3.4  
  & \cellcolor{green!20}+0.2  
  & \cellcolor{red!20}-1.6  
  & \cellcolor{red!20}-0.7  
  & \cellcolor{red!20}-2.4  
  & \cellcolor{green!20}+1.5  
  & \cellcolor{green!20}+6.1  
  & \cellcolor{red!20}-9.2  
  & \cellcolor{red!20}-2.7  \\

\texttt{small} (\%) 
  & \cellcolor{red!20}-3.9  
  & \cellcolor{red!20}-4.7  
  & \cellcolor{red!20}-1.9  
  & \cellcolor{red!20}-2.6  
  & \cellcolor{green!20}+3.4  
  & \cellcolor{green!20}+0.9  
  & \cellcolor{green!20}+4.9  
  & \cellcolor{green!20}+1.6  
  & \cellcolor{green!20}+4.9  \\

\texttt{medium} (\%) 
  & \cellcolor{green!20}+0.8  
  & \cellcolor{green!20}+4.8  
  & \cellcolor{green!20}+3.9  
  & \cellcolor{green!20}+1.2  
  & \cellcolor{green!20}+0.3  
  & \cellcolor{red!20}-0.3  
  & \cellcolor{green!20}+3.7  
  & \cellcolor{green!20}+5.0  
  & \cellcolor{green!20}+8.2  \\

\texttt{large} (\%) 
  & \cellcolor{green!20}+3.6  
  & \cellcolor{green!20}+4.4  
  & \cellcolor{red!20}-0.5  
  & \cellcolor{green!20}+4.2  
  & \cellcolor{green!20}+5.7  
  & \cellcolor{green!20}+1.3  
  & \cellcolor{green!20}+2.8  
  & \cellcolor{green!20}+3.4  
  & \cellcolor{green!20}+5.0  \\
\bottomrule
\end{tabular}}
\end{table}

\end{document}